\pdfoutput=1 

\documentclass[letterpaper, 10 pt, journal, twoside]{IEEEtran}
\IEEEoverridecommandlockouts

\usepackage{amssymb,amsthm}
\usepackage{amsmath,amsfonts}
\usepackage{algorithmic}
\usepackage{algorithm}
\usepackage{array}
\usepackage{balance}
\usepackage{bookmark}
\usepackage{booktabs}
\usepackage{calc}
\usepackage{cite}
\usepackage{etoolbox}
\usepackage{eucal}
\usepackage{float}
\usepackage{graphics}
\usepackage{graphicx}
\usepackage{hyperref}
\hypersetup{colorlinks=true}
\usepackage{mathptmx}
\usepackage{multirow}
\usepackage[normalem]{ulem}
\usepackage{pifont,hologo}
\usepackage{pdfpages}
\usepackage{siunitx}
\usepackage{stfloats}
\usepackage{textcomp}
\usepackage{url}
\usepackage{verbatim}
\usepackage{wrapfig}
\usepackage{xcolor}
\usepackage{xspace}

\DeclareSIUnit\pixel{px}

\newcommand{\yes}    {\textcolor{green}{\ding{52}}}

\def\eqref#1{Eq.~\ref{#1}}
\def\figref#1{Fig.~\ref{#1}}

\def\tabref#1{Tab.~\ref{#1}}

\makeatletter
\DeclareRobustCommand\onedot{\futurelet\@let@token\@onedot}
\def\@onedot{\ifx\@let@token.\else.\null\fi\xspace}
\def\eg{\emph{e.g}\onedot} 
\def\ie{\emph{i.e}\onedot}

\def\wrt{w.r.t\onedot} 
\def\etal{\emph{et~al}\onedot}
\def\etalcite#1{\etal~\cite{#1}}
\makeatother

\def\transpose{^\intercal}

\title{Motion-Aware Optical Camera Communication\\with Event Cameras}

\author{Hang Su, Ling Gao, Tao Liu, and Laurent Kneip%
\thanks{Manuscript received: June, 24, 2024; Revised October, 19, 2024; Accepted November 27, 2024. This letter was recommended for publication by Editor Hyungpil Moon upon evaluation of the Associate Editor and Reviewers' comments. We would like to acknowledge the funding support provided by project 62250610225 by the Natural Science Foundation of China, as well as projects 22DZ1201900, 22ZR1441300, and dfycbj-1 by the Natural Science Foundation of Shanghai.}
\thanks{The authors are with the Mobile Perception Lab of the School of Information Science and Technology, ShanghaiTech University (\protect\url{https://mpl.sist.shanghaitech.edu.cn}).}
\thanks{Digital Object Identifier (DOI): see top of this page.}
}

\begin{document}
\maketitle
\begin{figure*}[ht!]
  \centering
  \includegraphics[width=\linewidth, page=1]{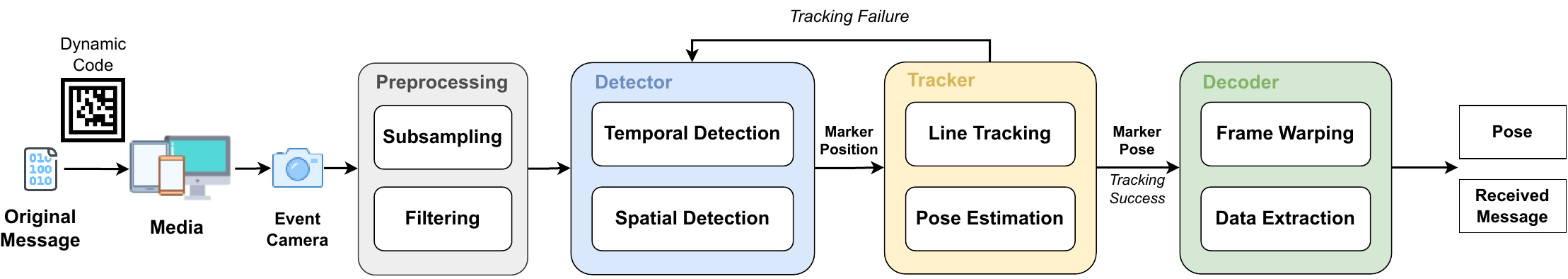}
  \caption{The overview of the proposed event-based optical camera communication system. The original message is encoded and displayed on media. The modulated visible lights are captured by the event camera as the receiver. Through our designed detector, tracker and decoder, the camera pose and received message can be recovered at the same time.}
  \label{fig:overview}
\end{figure*}
\begin{abstract}
As the ubiquity of smart mobile devices continues to rise, Optical Camera Communication systems have gained more attention as a solution for efficient and private data streaming. This system utilizes optical cameras to receive data from digital screens via visible light. Despite their promise, most of them are hindered by dynamic factors such as screen refreshing and rapid camera motion. CMOS cameras, often serving as the receivers, suffer from limited frame rates and motion-induced image blur, which degrade overall performance. To address these challenges, this paper unveils a novel system that utilizes event cameras. We introduce a dynamic visual marker and design event-based tracking algorithms to achieve fast localization and data streaming. Remarkably, the event camera's unique capabilities mitigate issues related to screen refresh rates and camera motion, enabling a high throughput of up to 114 Kbps in static conditions, and a 1 cm localization accuracy with 1\% bit error rate under various camera motions.
\end{abstract}


\section*{MULTIMEDIA MATERIAL}

The code implementation, along with the dataset and video, can be found at \protect\url{https://github.com/suhang99/EventOCC}

\section{INTRODUCTION}
\label{sec:intro}

\IEEEPARstart{T}{he} rapid advancement in communication technologies has driven the exploration of innovative data streaming methods. Optical Camera Communication (OCC) has emerged as a promising solution for leveraging optical cameras and digital screens to address issues inherent to traditional communication systems, such as interference, radio frequency spectrum congestion, and privacy concerns~\cite{matheus2019visible}. OCC systems employ screens to stream data that optical cameras capture and decode, utilizing visible light to reduce interference and enhancing data privacy through line-of-sight communication. Given the widespread presence of cameras on smartphones and digital screens in modern infrastructure, OCC offers a cost-effective and easily deployable solution. Its unique advantages make it suitable for various applications~\cite{matheus2019visible} including indoor navigation, augmented reality/virtual reality (AR/VR), and device-to-device private data streaming. 

Despite its numerous benefits, OCC systems face several challenges~\cite{saeed2019optical}. Ambient light can significantly hamper data rates and signal integrity. Additionally, camera motion and screen frame switching introduce image blur, which severely impacts the system’s ability to accurately locate and decode streamed data. Several efforts to address these issues have been made, including designing motion-resistant markers and algorithms~\cite{hao2012cobra, perli2010pixnet,nguyen2016high,li2015real} which allow for robust detection and data streaming under distortion and rapid camera motion. Other works~\cite{hu2013lightsync,saeed2019optical} focus on handling the synchronization between screens and cameras so that the captured images can be matched with the frames on the screen. However, none of these OCC systems perform well under rapid motion. The receiver in OCC systems, typically CMOS cameras, often becomes a bottleneck in achieving reliable performance. The limitations are primarily due to dynamic factors such as camera motion and screen refresh rates, which conventional imaging sensors struggle to handle effectively.

This paper presents a novel event-based OCC system to simultaneously achieve high data-rate streaming and accurate localization. Our primary contribution consists of employing an event camera, also known as a Dynamic Vision Sensor~\cite{gallego2020event}, in our OCC system. We also introduce a novel dynamic marker, compatible with any digital displays (\ie smartphones, screens, and tablets), to facilitate continuous data streaming and localization with event cameras. Our event-based OCC system is capable of performing reliable data streaming in highly dynamic scenarios where existing OCC systems usually fail due to motion blur. Event cameras are bio-inspired visual sensors that have gained popularity in recent years. Unlike regular cameras that generate images frame-by-frame, the pixels of an event camera act asynchronously and individually trigger events whenever the perceived intensity changes by more than a threshold amount. This unique imaging mechanism ensures high temporal resolution in the order of microseconds and low power consumption. These attributes make event cameras an ideal fit for OCC systems, addressing the limitations of conventional imaging sensors which are easily affected by dynamic scenes. 

The proposed event-based OCC system is able to perform continuous data streaming and accurate localization under rapid camera motion. Our experiments validate the effectiveness of our system in terms of data streaming throughput and localization accuracy. We showcase that our method is more reliable than frame-based OCC systems under rapid camera motion. We also demonstrate the potential of our event-based OCC system in a real-world AR scenario. Our detailed contributions are as follows:
\begin{itemize}
    \item We propose an optical camera communication system based on event cameras that supports continuous data streaming and accurate localization.
    \item We design a dynamic marker that is suitable for robust detection under camera motion, and its corresponding detection, tracking, and decoding algorithms.
    \item We conduct three experiments to demonstrate our proposed system: data streaming capacity, localization performance under dynamic motion, and a baseline comparison with traditional frame-based systems.
\end{itemize}

\section{RELATED WORKS}
\label{sec:related_works}


\subsection{Optical Camera Communication Systems}

With the exponentially growing number of digital displays and smart devices such as smartphones and tablets, Optical Camera Communication (OCC)---sometimes referred to as Visible Light Communication~\cite{pathak2015visible,matheus2019visible,saeed2019optical}---has attracted much interest in both academia and industry. In an OCC system, information is streamed through visible light from lighting devices such as screens to optical sensors. Various interesting applications in areas such as the Internet of Things, indoor localization, and AR/VR have been developed. The past decade has witnessed the emergence of numerous solutions. 

For instance, PixNet~\cite{perli2010pixnet} introduces spatial orthogonal frequency division multiplexing (OFDM) to encode information and applies a perspective correction algorithm to handle image distortion. COBRA~\cite{hao2012cobra} introduces a novel 2D color barcode that can be tracked by a corner detector. Building upon COBRA, LightSync~\cite{hu2013lightsync} achieves unsynchronized communication between LCDs and cameras, and doubles the throughput. Some other methods~\cite{nguyen2016high,li2015real,yuan2012dynamic} embed signals into regular images thereby achieving human-invisible data streaming. However, the low sampling rates and motion blur caused by CMOS cameras hinder efficient and reliable OCC applications. Although various remedies such as channel coding and synchronization have been proposed~\cite{pathak2015visible,saeed2019optical}, these issues remain challenging to address. Our event-based OCC method effectively and easily addresses challenges such as motion blur and frame synchronization while ensuring reliable data streaming.

Unlike CMOS cameras, event cameras can naturally detect illumination changes at very high rates and do not suffer from motion blur or frame synchronization. The potential of event cameras towards OCC has been demonstrated recently. Perez-Ramire~\etalcite{perez2019optical} designed an OCC system featuring an event camera and an LED tag capable of working in a static scene. Nakagawa~\etalcite{nakagawa2023multi,nakagawa2024linking} utilized RGB and event cameras with modulated LED lights in a multi-agent system to achieve visual coordination. However, the application of event camera in screen-based OCC system has not yet been explored. In our work, we design the first event-based OCC method that can work on ubiquitous displaying mediums.


\subsection{Event-based Detection and Tracking}

In recent years, a variety of algorithms have been proposed to parameterize asynchronous event streams into low-level features such as corners or lines \cite{vasco2016fast,mueggler2017fast,alzugaray2018asynchronous,scheerlinck2019asynchronous,chamorro2022event}. However, directly applying such feature tracking algorithms~\cite{zhu2017event, alzugaray2020haste, sanyal2024ev} is not always effective in an OCC system due to the constantly changing dynamic content. Instead, we incorporate salient line features in our proposed dynamic marker that can be easily distinguished. Event-based line trackers rely on various techniques such as the Hough transform~\cite{mueggler2014event}, non-parametric formulations~\cite{brandli2016elised}, and spatiotemporal formulations~\cite{everding2018low,dietsche2021powerline,gao2023eventail,gao2024eventail}. In particular, Mueggler~\etalcite{mueggler2014event} mounted an event camera on a highly agile drone and estimated its ego-motion by tracking a static, black square pattern. Everding~\etalcite{everding2018low} approximated the structure of the raw events in the space-time volume by planes and updated the line representations using PCA. Inspired by these works~\cite{mueggler2014event,dietsche2021powerline}, we have designed a new line tracker which is suitable for our task.

Marker-based localization methods for event cameras have recently also been explored. The first category involves marker-based tracking and decoding, where the transmitted messages are limited and immutable. In~\cite{sarmadi2021detection} the marker is tracked by line segments from event images while~\cite{loch2023event} updates the rotation and translation event-by-event. Hole{\v{s}}ovsk{\`y}~\etalcite{holevsovsky2021experimental} compared the ability of regular global shutter cameras and event cameras to detect markers under dynamic motion. The second category further leverages the sensor's high temporal resolution to handle high-frequency oscillating light sources (\eg~beacons) amid background noise or events caused by ego-motion. Censi~\etalcite{censi2013low} used a fixed event camera to estimate the drone's pose by tracking blinking LEDs at high frequency (above \SI{1}{\kHz}), achieving centimeter-level positioning accuracy. Chen~\etalcite{chen2021novel,chen2020novel} adopted a similar setup for accurate positioning and gesture tracking. Note that, while most LED-based markers can be considered dynamic, they communicate only very limited information (\eg~a fixed ID) and are difficult to install. In contrast, our proposed dynamic marker is designed specifically for digital display mediums, such as digital screens or smartphones, enabling fast and robust localization capability alongside customized data streaming usability. Furthermore, the marker is also tailored for event cameras, while other works primarily adapt existing markers for use with event cameras.

\section{DYNAMIC MARKER DESIGN}
\label{sec:marker_design}

Modern fiducial markers, such as AprilTag~\cite{krogius2019flexible}, ArUco~\cite{garrido2014automatic}, are designed to be easily distinguished from natural scenes. They typically consist of a payload and a locator, enabling robust detection under distortion and blurring. Despite their successful applications (\ie~SLAM, AR, OCC and camera calibration), these markers are not usually suitable for event cameras, particularly in our task. As depicted in~\figref{fig:code}, we have designed the first dynamic marker that varies over time to enable stable detection by an event camera. The detailed design is as follows.

The transmitted information (\ie~payload) is first converted into binary form and arranged in a square pattern. Each cell carries one bit of information, with black indicating $\mathtt{0}$ and white indicating $\mathtt{1}$. The choice of black and white ensures high contrast, thus stabilizing the triggering of events when changing the pattern. The \emph{effective payload} is located at the center of the marker and surrounded by a one-unit square ring named the \emph{interior locator}. This locator has two adjacent solid edges on one side and two adjacent dashed edges on the other side, an asymmetric arrangement used to determine the marker's orientation. The interior locator is enclosed by another three-layer square ring named the \emph{exterior locator}, which consists of a square ring of white cells on a black background. Note that the central effective payload and the interior locator are blinking constantly, while the exterior locator remains static.

The static exterior locator triggers events only when the camera is moving, and these events are utilized for the asynchronous update of the camera pose. The design of the exterior locator ensures continuous event triggering during camera motion, allowing it to be easily detected and tracked in both space and time. This, in turn, enables stable camera pose estimation.

To stream information continuously, we employ a switching scheme that alternates between data frames (comprising all three components) and the blank frames (featuring only the exterior locator) at a fixed, pre-set frequency that is consistent with the digital display. This scheme makes the triggered events from each data frame independent of previous data frames, ensuring that errors from previous frames do not accumulate into the current frame.

\definecolor{marker_green}{HTML}{2F8C16}
\definecolor{marker_purple}{HTML}{282FAB}
\definecolor{marker_teal}{HTML}{38758C}
\begin{figure}[t]
  \centering
  \includegraphics[width=0.95\linewidth]{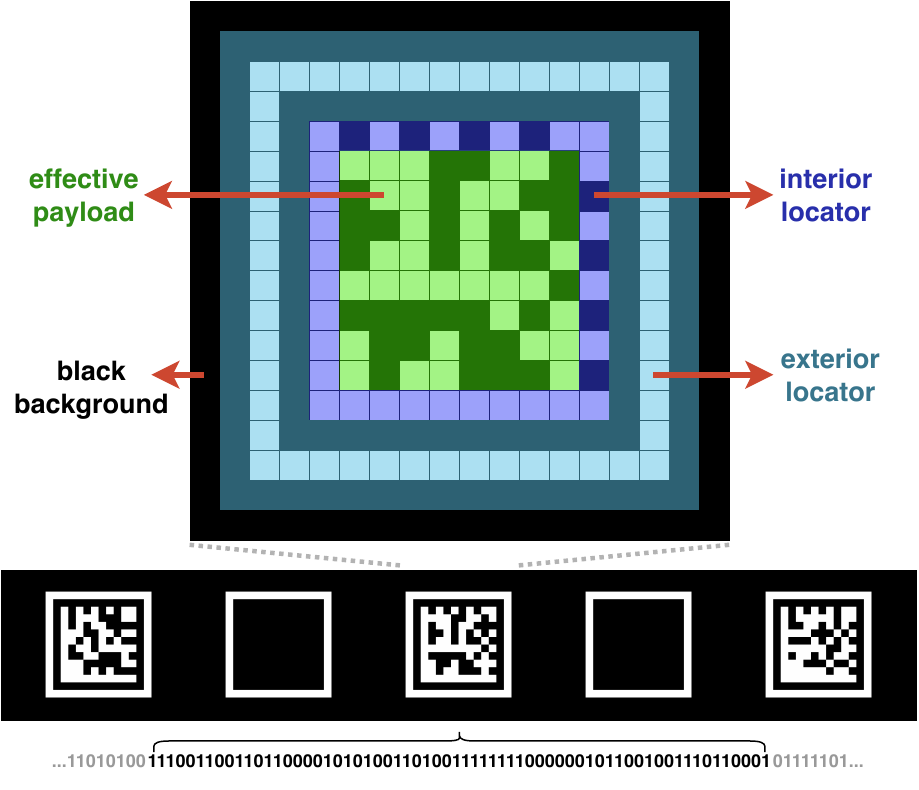}
  \caption{Dynamic Marker Design. The payload cells (in \textbf{\textcolor{marker_green}{green}}) carry the transmitted information. Cells marked by \textbf{\textcolor{marker_purple}{purple}} and \textbf{\textcolor{marker_teal}{teal}} serve as the interior and exterior locators, respectively. We alternately display data frames and blank frames on a \textbf{\textcolor{black}{black}} background like a slide show. Note that the effective payload and the interior locator are dynamic, while the exterior locator remains static. Colors and each cell's boundary are used only for clarity.}
  \label{fig:code}
\end{figure}

\begin{figure*}[t]
    \centering
    \includegraphics[width=\linewidth]{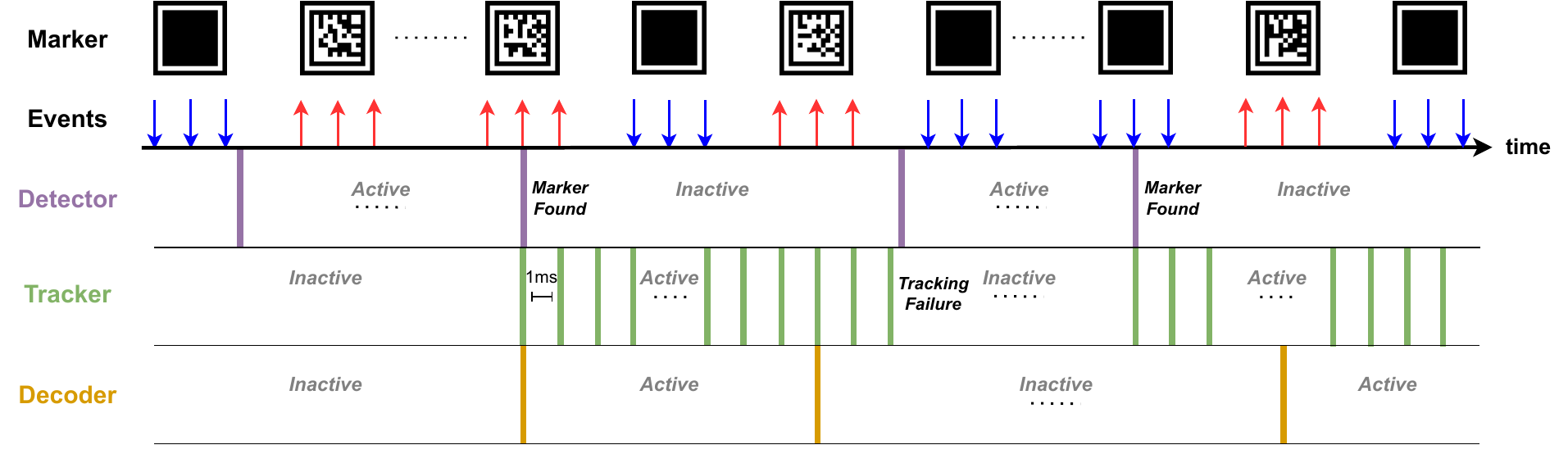}
    \caption{Schematic diagram of the process of detector, tracker and decoder. Blue arrows stand for negative events while red ones are positive events.}
    \label{fig:timeline}
\end{figure*}

\section{SYSTEM DESIGN}
\label{sec:system_design}

The proposed system, outlined in~\figref{fig:overview}, is designed to effectively and efficiently detect, track, and decode the aforementioned dynamic marker with an event camera over time. The transmitted messages, encoded into a sequence of binary patterns, are broadcast on a digital display medium at a fixed frequency. An event camera then captures this information through the dynamic marker. To process this event stream, the system begins with preprocessing techniques to suppress background noise and then proceeds with three key modules sequentially.

As shown in~\figref{fig:timeline}, the \emph{detector} works at a relatively low frequency (\eg~\SI{1}{\Hz}), typically serving as the bootstrapping point or for refining processes. It searches for the periodic and structured components in the event stream to spatially and temporally localize the marker. In contrast, the \emph{tracker} operates at a high frequency (\eg~\SI{1}{\kHz}), continuously maintaining the tracking under various motion. Lastly, the \emph{decoder} functions at the same, pre-determined frequency as the data frame. This module is responsible for recovering the data embedded in the localized effective payload of the marker. Finally, we can obtain both the marker's location and streamed data for various applications.


\subsection{Pre-Processing Module}

An event camera outputs a series of asynchronous events responding to temporal logarithmic brightness changes. Each individual event $e \doteq (u, v, t, pol)$ is represented by its pixel location $\mathbf{p}_e \doteq (u, v)\transpose \in \mathbb{R}^2$, its timestamp $t$, and its polarity $pol \in \{-1, +1\}$ (\ie~a single bit indicating whether the brightness increased or decreased). Besides processing raw events, we also employ a popular data representation called the \emph{Surface of Active Events}~(SAE)~\cite{benosman2013event}. The SAE is a function $\mathcal{S}: \mathbb{N}^2 \rightarrow \mathbb{R}^+$ that maps a pixel coordinate to its most recent event's timestamp, \ie~$t = \mathcal{S}(\mathbf{\mathbf{p}_e})$. The SAE is an asynchronously updated data structure building the bridge to traditional computer vision methods.

We add two sequential filters~\cite{everding2018low} to the raw events to pick out proper candidates for downstream tasks. Refractory filters applied at each pixel suppress the burst of events induced by sudden brightness changes, which is particularly common in our task given the periodic changes between black and white. After one event has been recorded, any other event at the same pixel location within a small time interval (\eg~\SI{1}{\ms}) will be discarded. Next, for every recorded event we check if at least three more events have also been recorded within a small spatiotemporal neighborhood (\eg~\SI{3}{pix}$\times$\SI{3}{pix}$\times$\SI{10}{\ms}). Only events that pass both checks will be used for further processing. 


\subsection{Detection Module}

\subsubsection{Temporal Detection}

A large amount of same-polarity events is periodically triggered when switching from blank frames to data frames and vice versa. In the first case, lots of cells are switching from black to white, thus causing a flood of positive events. In the second case, switching from white to black causes abundant negative events. The periodic event floods can be observed in the distribution of the number of events over time. The latter will indeed resemble a pulse wave. If we are able to estimate the phase of the periodic pulse, we will know precisely when the digital display medium has been refreshed.

We follow the idea in~\cite{censi2013low} to search for a region where events are periodically triggered. Basically, we first search for all event triplets triggered consecutively at the same pixel, with adjacent events having opposite polarities, \ie~``pos-neg-pos'' or ``neg-pos-neg''~\cite{censi2013low}. The duration of each event triplet is the timestamp difference between the first and last event. Then we assign a Gaussian weight to each event triplet, where the mean is the difference between its duration and the frequency of data frames. By accumulating the weight of each pixel, we can construct a heatmap of the image canvas which indicates the probability of the marker's presence. After thresholding and extracting the contours of the heatmap, we obtain multiple Regions of Interest (ROI) that potentially show the dynamic marker. We simply choose the ROI with the largest area as the target.

The periodically triggered events can be located by Fourier analysis in the temporal domain. With the identified ROI, we convert the selected events into a discrete signal from the number of events within each small time interval. We segment a chosen period of time (\eg~\SI{1}{\s}) into $N$ equal intervals. The duration of these sub-intervals is determined by a sampling rate, which is set to be at least twice the data frame frequency $f_d$, in accordance with the Nyquist-Shannon sampling theorem. For each interval, we count the number of positive events\footnote{Either positive or negative events can be applied here to estimate the phase, of which the only difference is a shift of \SI{180}{\degree}. within the identified ROI. Note that the frequency of the data frames $f_d$ equals the frequency at which positive events are triggered by the marker. The event count forms the values $x_n$ of the discrete signal. We apply the Discrete Fourier Transform (DFT) to obtain the refreshing time $\tau_m$ of the $m$-th data frame from the start of data streaming by
\begin{equation} 
    X_k = \sum_{n=0}^{N-1}x_n \exp(-\frac{i \cdot 2\pi}{N}kn) \quad k = 0,\ \ldots,\ N-1 \,,
\end{equation}
\begin{equation}
    \tau_m = \frac{2\pi - \arg(X_k)}{2\pi f_d} + \frac{m}{f_d}\quad m \in \mathbb{N} \,,
\end{equation}
where $arg$ takes the phase of the discrete signal.
}

\subsubsection{Spatial Detection}

Once the dynamic marker has been temporally detected, we can refine its spatial location, where the payload with the interior locator is located. We analyze the SAE around the data frame arrival time $\tau$. To make it easier to process, we first normalize the SAE, denoted as $\hat{\mathcal{S}}$,
\begin{equation}
    \hat{\mathcal{S}}(\mathbf{p}) = \left\{
    \begin{aligned}
      & \ \ \ \ \ 0 & 
      \text{ if } \mathcal{S}(\mathbf{p}) \notin [\tau - \Delta t, \tau + \Delta t] \\
      & \frac{\mathcal{S}(\mathbf{p})-(\tau - \Delta t)}{2\Delta t} &
      \text{ if } \mathcal{S}(\mathbf{p}) \in [\tau - \Delta t, \tau + \Delta t]
    \end{aligned}
    \right. \,,
    \label{eq:normalized_sae}
\end{equation}
where $\tau$ is the estimated data frame arrival time from the temporal detection module and $\Delta t$ is a small time interval (practically set to $0.2$ of the period $\frac{1}{f_d}$). 

Next, we perform contour extraction on the normalized SAE $\hat{\mathcal{S}}$ to identify all closed elements. We approximate these elements into convex polygons, and we then filter this collection using three specific criteria, polygons must \emph{(i)} consist of at least four vertices, \emph{(ii)} cover a significant area (e.g. larger than a square with \SI{50}{pix} length), and \emph{(iii)} feature edges that are almost perpendicular to each other with roughly equivalent lengths. Our marker is uniquely designed to meet these criteria. Specifically, the left and bottom parts of the interior locator construct a distinct ``L''-shape structure. This design makes the marker easily distinguishable from other shapes and ensures it will be identified from the set of candidate polygons. The ``L''-shape structure also helps to determine the orientation of the marker.


\subsection{Tracking Module}

\subsubsection{Edge Tracking}

Given that the data payload is unknown and changing, it is challenging to locate the marker by tracking features from the dynamic region, especially during the blank time with rapid camera motion. Instead, we choose to track the salient edges of the exterior locator by asynchronously updating their location as nearby events are triggered.

We focus on tracking the eight salient edges, created by the gradient between the black and white layers of the exterior locator of the marker. Inspired by \cite{mueggler2014event, dietsche2021powerline}, each edge segment $\ell$ is manifested by a cluster of nearby events. For each incoming event at location $\mathbf{p_e}$, we compute the orthogonal distance $\operatorname{dist}(\mathbf{p}_e, \ell)$ between each edge segment and event.\footnote{Note that if the orthogonal projection is not on the edge segment, we calculate the Euclidean distance to its nearest endpoint.} If the distance is below a certain threshold $\theta_d$, the event will be added to the cluster of that edge. To prevent stale events from corrupting the edge, we set a maximum capacity for each cluster. A small capacity may lead to tracking instability, while a big one results in high latency. To balance both latency and accuracy, we set the capacity to be positively correlated with the length of each edge segment, \ie~$|\mathcal{E}_\ell| \doteq \alpha \cdot \operatorname{length}(\ell)$, where $\mathcal{E}_\ell$ denotes a set of events representing the cluster of the edge and $\alpha$ is a scaling factor. We manually set $\theta_d = \SI{6}{pix}$ and $\alpha = 0.3$.

As indicated in~\cite{mueggler2014event}, the FIFO event replacement policy may fail due to camera rotation. To ensure that events in the cluster are evenly distributed around the edge segment, we divide the edge segment evenly into $|\mathcal{E}_\ell|$ parts. Each part keeps track of a single, recent, and sufficiently close event. For each incoming event, we replace the event whose projection lies in the same segment part.

\subsubsection{Pose Estimation}

Next, the update of the camera pose is performed through nonlinear optimization. We define a world coordinate frame attached to the marker, assuming it is statically placed in the environment. The origin is located at the top left corner, the \emph{x}-axis follows the top horizontal line, the \emph{y}-axis follows the left vertical line, and the \emph{z}-axis points away from the camera. Let $\mathbf{P} \in \mathbb{R}^3$ be the location of a 3D point in the world coordinate frame. The projected point $\mathbf{p}$ onto the image plane is given by
\begin{equation}
  \mathbf{p} = \pi(\mathbf{P}) = \mathbf{K}(\mathbf{R} \mathbf{P} + \mathbf{t}) \,,
\end{equation}
where $\mathbf{K}$ is the camera intrinsic matrix, $\mathbf{R}$ is the rotation matrix, and $\mathbf{t}$ is the translational vector. Given the known position of all the exterior locator's corners, we can derive the pixel coordinates of the endpoints of each edge segment by projection, \ie~$\ell \doteq \pi(\mathbf{L})$. The camera pose is then estimated by minimizing the sum of distances between the events in $\mathcal{E}_\ell$ and its corresponding projected edge segment as in
\begin{equation}
  \mathbf{R}^*, \mathbf{t}^* = \mathop{\operatorname{argmin}} \limits_{\mathbf{R}, \mathbf{t}}
                               \sum_{i=1}^8 \sum_{e}^{\mathcal{E}_{\ell_i}} 
                               \operatorname{dist}(\mathbf{p}_e, \pi(\mathbf{L}_i)) \,.
\end{equation}
The minimization is performed by the Levenberg-Marquardt method, with a robust kernel function such as Huber norm.

\subsubsection{Update Rule}
Higher tracking frequency allows for lower latency and adaptation to varying speeds, but leads to more computational cost for pose estimation. Instead, we update the pose asynchronously by using a tracking latency indicator which is the mean distance between each edge segment and its corresponding event cluster. A large distance typically indicates that the current pose has become mismatched, preventing new incoming events from being assigned to the cluster. This signals the need for a pose update. With a negligibly small overhead, this indicator calculation operates alongside the tracking module. The costly pose estimation runs asynchronously only when more than one indicator is greater than $0.5 \cdot \theta_d$.


\subsection{Data Decoding Module}

The data decoding module is activated every time a new data frame comes in. Considering the refresh time of common digital display mediums, we define a small time interval $\Delta \tau$ around $\tau$, to find the positive events\footnote{Given the switching scheme between data and blank frames, either positive or negative events are sufficient for data decoding. We opt to use the positive events and thereby process information right after initial reception.} that were triggered by the data frame. We use~\eqref{eq:normalized_sae} to truncate and normalize the SAE. Next, we calculate the homography that unwarps the SAE onto a fixed template location in the image. This unwarped SAE, referred to as $\hat{\mathcal{S}}_{\mathbf{H}}$, is further binarized into an image $\mathcal{B}$, that is
\begin{equation}
  \mathcal{B}(\mathbf{p}) = \left\{
  \begin{aligned}
    \mathtt{1} \text{ if } \hat{\mathcal{S}}_{\mathbf{H}}(\mathbf{p}) >   \theta_b \\
    \mathtt{0} \text{ if } \hat{\mathcal{S}}_{\mathbf{H}}(\mathbf{p}) \le \theta_b
  \end{aligned}
  \right. \,,
\end{equation}
where $\theta_b$ is the threshold used for binarization. A median filter of size $5$ is also applied to $\mathcal{B}$ for a smoother image.

After these operations, each cell of the payload can be easily located in $\mathcal{B}$. Let $s_c$ be the length of a cell in pixels. The bit $c_{ij}$ is carried by the cell in the $i$-th row and the $j$-th column and can be extracted using
\begin{equation}
  c_{ij} = \left\{
  \begin{aligned}
    \mathtt{1} \text{ if } \sum \mathcal{B}[i \cdot s_c, j \cdot s_c, s_c, s_c] >   0.5 \cdot s_c^2 \\
    \mathtt{0} \text{ if } \sum \mathcal{B}[i \cdot s_c, j \cdot s_c, s_c, s_c] \le 0.5 \cdot s_c^2
  \end{aligned}
  \right. \,,
\end{equation}
where $\mathcal{B}[a, b, c, d]$ stands for a region of $\mathcal{B}$ starting at row $a$ and column $b$ with $c$ rows and $d$ columns. At least half of the cell area needs to be $\mathtt{1}$ in order for $c_{ij}$ to become $\mathtt{1}$.

Most digital screens operate in a progressive scanning manner. Expanding the time window $\Delta \tau$ can ensure that all payload cells have been refreshed. However, a large $\Delta \tau$ may cause neighboring cells to affect each other under rapid motion. In practice, the decoding process is divided into multiple steps, with each step addressing a successive segment of the overall time window. In each step, we use the most recent tracking results for decoding. Each step only covers partial information $c^k$ which is decoded from the recently refreshed part of the screen. The complete information for each bit $c_{ij}$ can be recovered by taking the logical \emph{OR} operation over all partial information, \ie~$c_{ij} \doteq c^1_{ij} \lor c^2_{ij} \lor \cdots \lor c^n_{ij}$.

\section{REAL-WORLD EXPERIMENTS}
\label{sec:experiment}

In order to validate the practicality of our event-based OCC system, we conduct three sets of experiments. First, we focus on the data streaming performance in terms of throughput and bit error rate in a stationary scenario, while considering external factors like the density of the marker's payload and the level of ambient light. Secondly, we demonstrate the joint ability to perform accurate localization and data streaming under rapid camera motion. This is further exemplified by a simple AR application in real-world scenarios. Thirdly, we compare our system against a QR code-based baseline method, which relies on a normal CMOS camera. We evaluate data streaming success rate under various levels of camera motion.


\subsection{Hardware Description}

We use a Prophesee Gen4 CD event camera (1280$\times$720 resolution) with Kowa LM5JCM lens (\SI{1033}{pix} focal length with an aperture of f/8) as the receiver and a DELL U2417H LCD monitor (\SI{60}{\Hz} refresh rate) as the transmitter. For the comparison experiment, we use the FLIR Blackfly S RGB camera at an acquisition rate up to 120Hz. The ground truth of the camera pose in the localization experiment is recorded by OptiTrack motion capture system.


\subsection{Data Streaming Capability}

\subsubsection{Experiment Setup}

The event camera is placed statically \SI{1.25}{\m} in front of the digital display medium such that the marker practically fills the entire view of the camera. We randomly generate binary sequences as the data payload. Each binary sequence is encoded in the form of the proposed dynamic marker and converted into a \SI{60}{fps} video which is compatible with the digital display medium. To find out the maximum achievable throughput, we test a series of markers that differ in the information density of the payload. Each cell is the smallest square unit on the marker, responsible for conveying one bit of information. The number of cells along one dimension is varied from 16 to 96 in the step of 16. To study the influence of ambient light, we also repeat the experiment in three different illumination conditions (\SI{10}{\lux}, \SI{250}{\lux}, and \SI{1000}{\lux}), which correspond to nighttime lighting, normal indoor lighting, and indoor with sunlight. Under the same experimental conditions (payload and lighting), we repeat the streaming ten times (each time sending a different binary sequence) and then average the results. Each data streaming takes around \SI{15}{\s}. We use the commonly used Reed-Solomon codes~\cite{wicker1999reed} for error correction. We investigate a collection of popular parameter choices of RS codes and apply the most suitable one for each payload size. The more bits used for error correction, the higher the probability to recover the entire information.\footnote{We select rs(255,251), rs(255,247), rs(255,239), rs(255,223), rs(255,159) and rs(255,127) for the aforementioned payload sizes, respectively. rs(\emph{n},\emph{k}) stands for a Reed-Solomon code of \emph{n} block length and \emph{k} message length.}

\begin{figure}
  \centering
  \includegraphics[width=\linewidth]{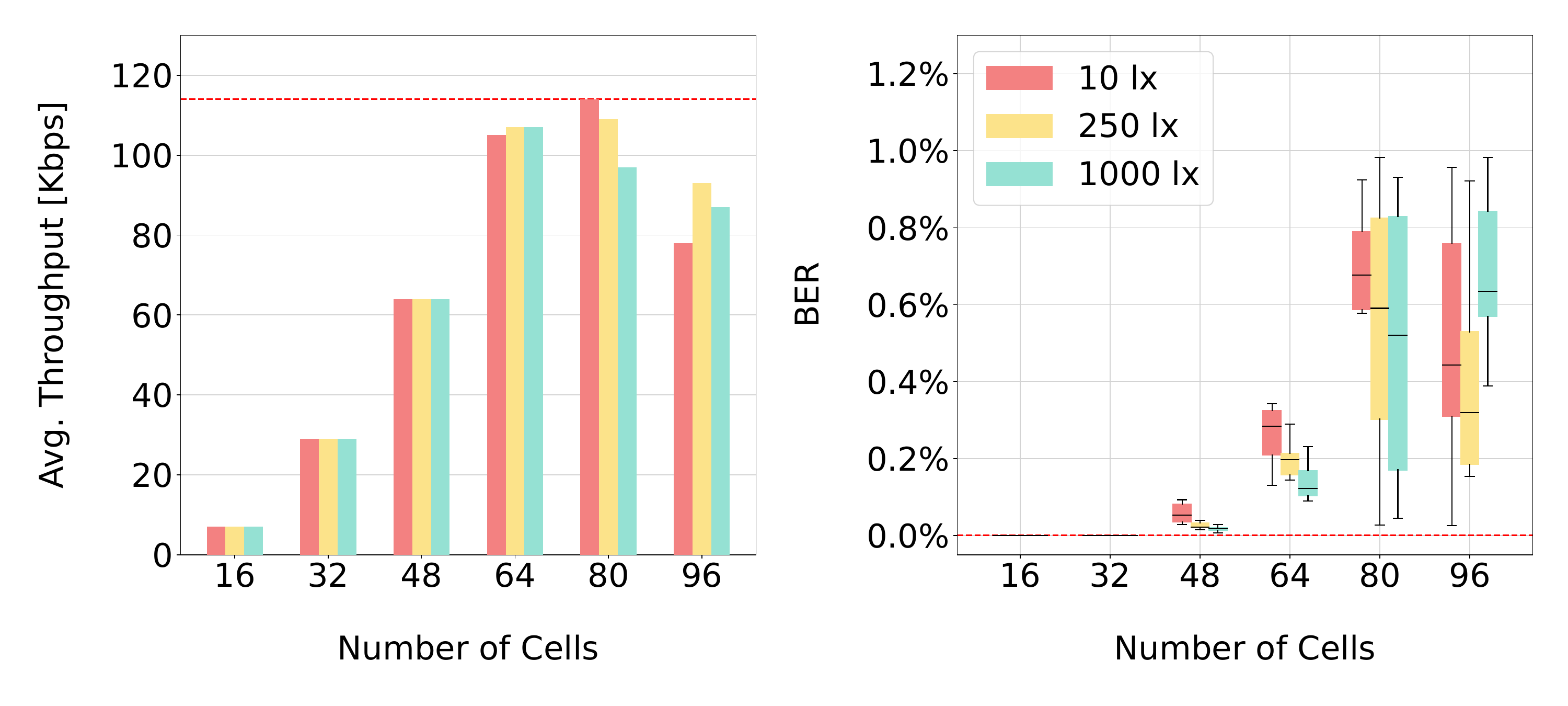}
  \vspace{-1em}
  \caption{The horizontal axis is the payload size on the marker and the vertical axis illustrates the two evaluation metrics. Different colors correspond to various lighting conditions. The box plot outlines the range between the first quartile to the third, where the median is marked by a line.}
  \label{fig:exp_a}
  \vspace{-1em}
\end{figure}

\subsubsection{Evaluation Metric}

We evaluate the data streaming capability in terms of \emph{Bit Error Rate} (BER) and \emph{throughput}. BER denotes the ratio between the number of error bits from received data and all transmitted bits before error correction. The throughput is the number of correctly decoded bits per second after error correction. Note that the error correction bits of the RS codes are not included in the throughput computation.

\subsubsection{Results Analysis}

As the payload cells become smaller and denser, the perceived pixels that cover each payload cell on the event camera naturally decrease. The choices of the payload size in~\figref{fig:exp_a} correspond to 30, 18, 13, 10, 8, and \SI{7}{pix} in the camera view. The maximum data streaming throughput is achieved at the payload size of 80$\times$80, and amounts to \SI{114}{Kbps}. From the results, we did not observe a significant influence of ambient light. In particular, for smaller payload sizes, all bits are correctly received, resulting in the same throughput and zero BER under all tested illumination conditions.

Since the overall BER remains small, the added error correction mechanism can thus fully recover the complete information in practically all cases. However, the BER increases as the payload size becomes larger because the pixel space for each payload cell is smaller and thus more easily affected by noise. When the payload size is larger than 96$\times$96, the BER will exceed \SI{1}{\percent} which requires a large proportion of the payload to be error correction bits. A low signal-to-noise ratio and a high ratio of error correction eventually result in a significant drop in throughput.

Additionally, we have performed experiments on the distance between the digital display medium with the dynamic marker and the event camera. With similar performance trends as in~\figref{fig:exp_a}, we found that the impact of both distance and payload size on throughput can be aptly represented by the number of pixels per cell.


\subsection{Localization Performance under Dynamic Motion}

\begin{table}[t]
  \centering
  \caption{Localization Accuracy}
  \vspace{-1em}
  \begin{center}
  \resizebox{\columnwidth}{!}{
  \begin{tabular}{ccccccc}
    \toprule
     & Time[s] & $\mathbf{R}_{\text{rpe}}$[$^{\circ}$]
               & $\mathbf{t}_{\text{rpe}}$[cm]
               & $\mathbf{t}_{\text{ate}}$[cm] 
               & BER & Decodable \\ 
    \midrule
    \emph{Seq. 1} & 107 & 0.29 & 0.53 & 1.55 & 0.78\% & \yes \\
    \emph{Seq. 2} & 112 & 0.32 & 0.57 & 1.40 & 1.18\% & \yes \\
    \emph{Seq. 3} & 104 & 0.23 & 0.38 & 1.43 & 0.72\% & \yes \\
    \emph{Seq. 4} & 112 & 0.28 & 0.52 & 1.63 & 1.04\% & \yes \\
    \emph{Seq. 5} & 110 & 0.33 & 0.66 & 1.96 & 1.13\% & \yes \\
    \bottomrule
  \end{tabular}}
  \end{center}
  \label{tab:localization}
  \vspace{-1em}
\end{table}
\begin{figure}
    \centering
    \includegraphics[width=0.95\linewidth]{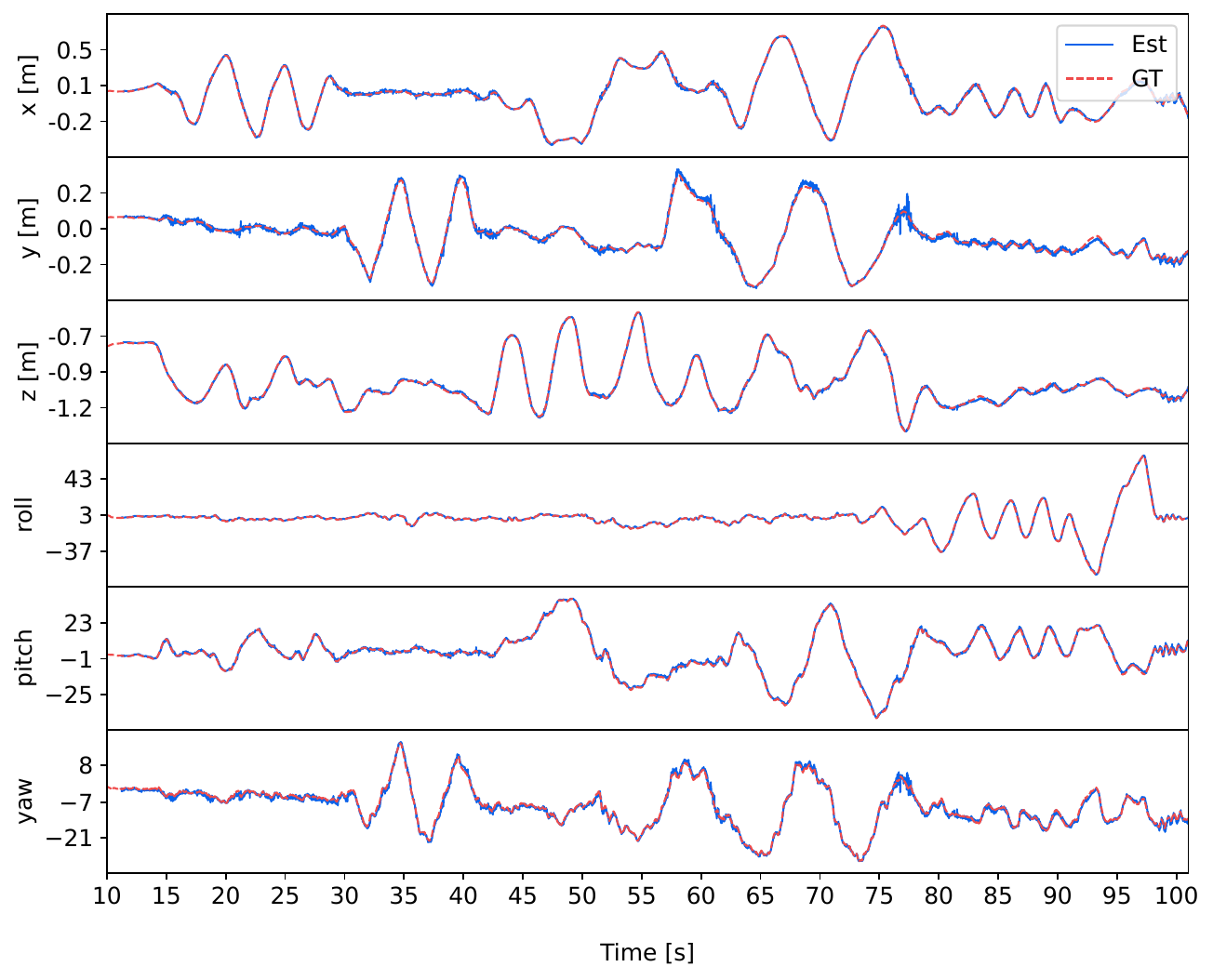}
    \vspace{-1em}
    \caption{Comparison between estimated camera poses and ground truth for the trajectory of \emph{seq. 3}.}
    \label{fig:localization}
    \vspace{-1em}
\end{figure}

\subsubsection{Experiment Setup}

We now restrict the payload to 16$\times16$ to increase the signal-to-noise ratio in each payload cell. We move the event camera, agilely and widely, while always keeping the marker in sight. The dynamic marker is running in full-screen mode against a pure black background on the screen of size \SI{27.2}{\cm}). The ground truth pose is captured at \SI{360}{\Hz} and furthermore aligned to be expressed \wrt the marker reference frame~\cite{gao2022vector}. An infrared filter is mounted in front of the camera to reduce the influence of the blinking LEDs of the motion capture system.

We have recorded five data sequences under normal indoor lighting with random information (except for the AR data sequence). Each data sequence covers multiple hand-held motion patterns and lasts about \SI{100}{\s}. The event camera is placed stationary for a few seconds after activating the tracking system so that we can synchronize the ground truth and estimated trajectory by aligning the time when both the ground truth pose and the number of events undergo a sudden change. We furthermore align our estimated trajectory and ground truth using Umeyama's method~\cite{umeyama1991least}, and fix the scale given the real-world size of the dynamic marker.

\subsubsection{Results Analysis}

We adopt the standard Relative Pose Error~(RPE) and Absolute Trajectory Error~(ATE)\footnote{\protect\url{https://github.com/MichaelGrupp/evo}} for evaluating the pose estimation. As shown in~\tabref{tab:localization}, the per-trajectory ATE error is ranging from \SI{1.40}{\cm} to \SI{1.96}{\cm} with a standard deviation of \SI{0.87}{\cm} to \SI{1.42}{\cm}. Accurate pose tracking is crucial for subsequent data decoding, where the aforementioned localization performance proves to be adequate to ensure successful data decoding (with error correction) among all data sequences. The average bit error rate of the overall streaming is around \SI{1}{\percent} and the data rate capacity is up to \SI{7.68}{Kbps} in the dynamic cases.

\textbf{AR demo}: We kindly refer the reader to our supplemental video for further qualitative results, demonstrating the joint ability to perform accurate localization and data streaming through an AR application. We believe our OCC system has the potential to enhance the performance of AR devices.


\subsection{Baseline Comparison}

\begin{figure}
    \centering
    \includegraphics[width=0.95\linewidth]{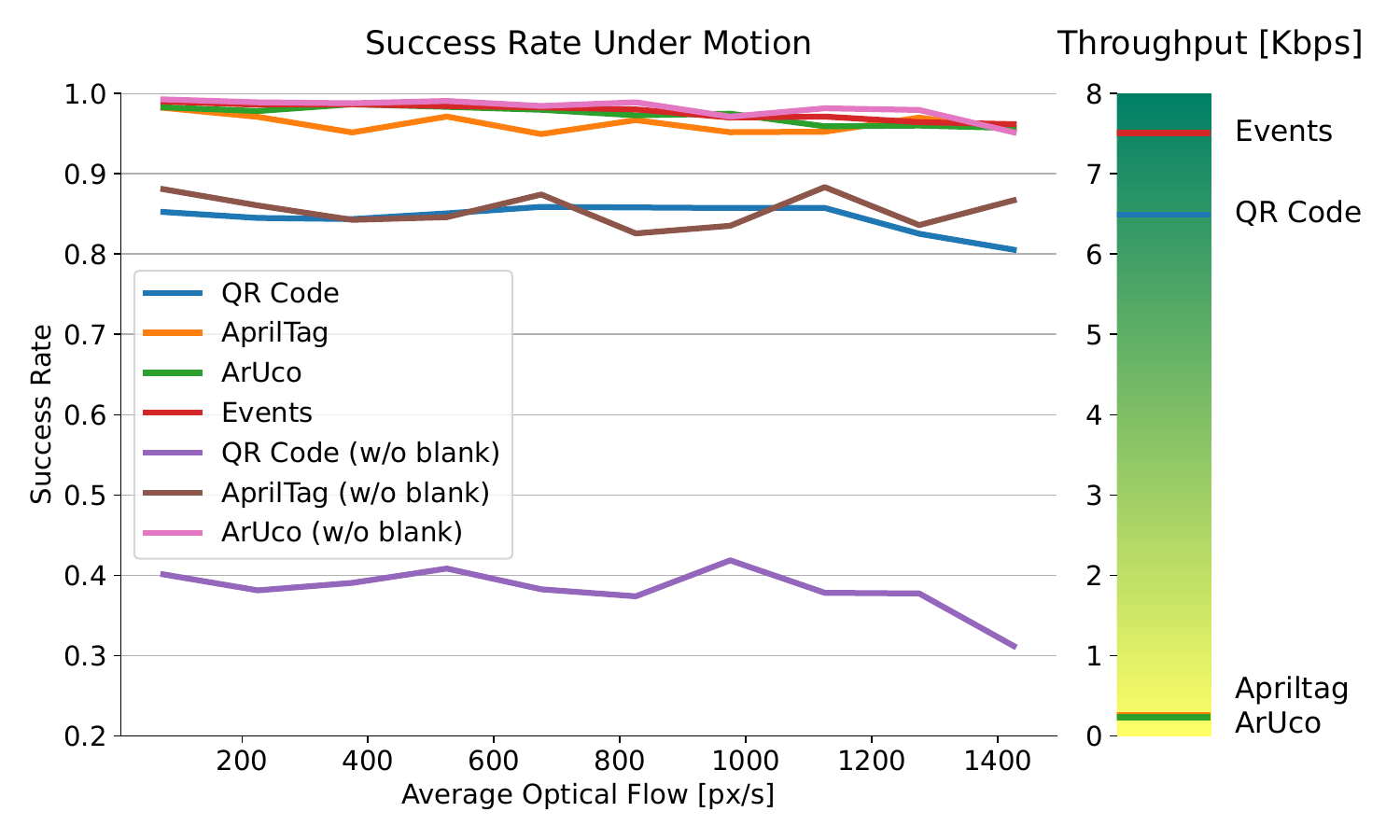}
    \caption{Comparison with frame-based baseline methods. Left: Data streaming success rate on different levels of optical flow. Right: Throughput for different systems.}
    \label{fig:comparison}
\end{figure}

We also conduct a comparison experiment to demonstrate the advantages of our event-based solution against frame-based OCC methods. Although various traditional frame-based OCC markers and algorithms~\cite{hu2013lightsync,hao2012cobra} have been designed, very few of them can support continuous data streaming under fast camera movement. We use a moving grayscale camera with QR codes, AprilTags, and ArUco markers as baseline methods for frame-based comparisons.

\subsubsection{Experiment Setup} We first crop the images to match the event camera's resolution. We use the QR Code (version 1) so that it has the most similar cell density with our designed marker with 16$\times$16 cells for payload. This version of the QR Code can also achieve the best performance in this OCC task with camera motion. We choose the tag36h11 version for AprilTag, and the $6\times6$ version for ArUco markers. The camera acquisition rate is set to be \SI{120}{\Hz}, and the screen refresh rate is \SI{60}{\Hz}.

\subsubsection{Evaluation Metric}

Instead of using absolute camera velocity, we evaluate performance based on the average optical flow of the marker in the image plane. This approach better accounts for the relative distance between the marker and the camera and challenges like motion blur.

\subsubsection{Results Analysis}

We compare our method with baseline solutions including QR codes, AprilTags, and ArUco markers. QR codes are typically used for transmitting large data payloads, while AprilTags and ArUco focus on robust tracking with lower data capacity. To facilitate these methods for OCC, we create two data streams: one continuously streaming codes or markers, and another alternating with blank frames. As shown in~\figref{fig:comparison}, our method performs similarly to AprilTags and ArUco, achieving nearly \SI{100}{\percent} success under camera motion. QR codes, however, achieve only \SI{85}{\percent} success with blank frames and \SI{40}{\percent} without, due to poor synchronization causing motion blur. The improvement with blank frames highlights their role in reducing interference between consecutive frames. We observe a drop in success rate for baseline methods as camera speed increases, mainly due to motion blur. Our event-based system, shown on the right side of~\figref{fig:comparison}, significantly improves throughput over baseline methods, and can be further enhanced with more cells as showcased in previous experiments. This concludes that our system provides both accurate tracking under motion and high throughput data streaming.

\section{CONCLUSION}
\label{sec:conclusion}

This paper introduces a novel OCC system for accurate localization and continuous data streaming. Our novelty consists of employing event cameras to unlock accurate localization in parallel with decent data rates under challenging conditions. Our efforts mark a pivotal step towards the next generation of efficient and reliable OCC systems. One of the limitations is the refresh rate of the standard digital displays, which is in stark contrast to the event camera's high temporal resolution (in the order of microseconds). We believe this technology to be highly promising, and are currently scaling the data streaming capacity by at least one order of magnitude through the adoption of faster visualization media like LED lights.

{

\bibliographystyle{bibtex/IEEEtran}
}

\end{document}